\documentclass[letterpaper]{article}
\RequirePackage{latexrelease}
\usepackage{natbib,alifeconf}  
\usepackage{siunitx,booktabs}
\usepackage{multirow}
\usepackage{algorithm}
\usepackage{algorithmic}

\title{Can Bio-Inspired Swarm Algorithms Scale to Modern Societal Problems?}
\author{Darren M. Chitty, Elizabeth Wanner, Rakhi Parmar \and Peter R. Lewis \\
\mbox{}\\
Aston Lab for Intelligent Collectives Engineering (ALICE) \\
Aston University, Aston Triangle, Birmingham, B4 7ET, UK\\
darrenchitty@googlemail.com} 

\begin{document}
\maketitle
\begin{abstract}\vspace{-0.2cm}
Taking inspiration from nature for meta-heuristics has proven
popular and relatively successful. Many are inspired by the
collective intelligence exhibited by insects, fish and birds.
However, there is a question over their scalability to the types
of complex problems experienced in the modern world. Natural
systems evolved to solve simpler problems effectively, replicating
these processes for complex problems may suffer from
inefficiencies. Several causal factors can impact scalability;
computational complexity, memory requirements or pure problem
intractability. Supporting evidence is provided using a case study
in Ant Colony Optimisation (ACO) regards tackling increasingly
complex real-world fleet optimisation problems. This paper
hypothesizes that contrary to common intuition, bio-inspired
collective intelligence techniques by their very nature exhibit
poor scalability in cases of high dimensionality when large
degrees of decision making are required. Facilitating scaling of
bio-inspired algorithms necessitates reducing this decision
making. To support this hypothesis, an enhanced Partial-ACO
technique is presented which effectively reduces ant decision
making. Reducing the decision making required by ants by up to
90\% results in markedly improved effectiveness and reduced
runtimes for increasingly complex fleet optimisation problems.
Reductions in traversal timings of 40-50\% are achieved for
problems with up to 45 vehicles and 437 jobs.
\end{abstract}

\vspace{-0.5cm}
\section{Introduction}
The natural world is filled with a wealth of differing animals and
ecosystems. Many of these organisms display collective behaviours
which they use to overcome problems within their ecosystem such as
ants foraging for food or bees communicating locations of nectar.
These organisms have inspired many computing algorithms to assist
in solving difficult real-world problems. Much of this inspiration
comes from the exhibition of collective behaviours whereby
thousands of organisms work together for the benefit of a colony,
flock or hive. Each organism is simplistic in nature and by itself
cannot survive but as part of a collective, problems such as
finding sources of food can be solved. Nature has been used as a
source of inspiration for the direct design of meta-heuristic
algorithms that are moderately successful in solving optimisation
problems of human consideration such as routing problems,
information management and logistics to name a few. Examples of
bio-inspired collective behaviour algorithms include Ant Colony
Optimisation (ACO) \citep{dorigo:1997} inspired by how ants forage
for food; Artificial Bee Colony (ABC) \citep{karaboga:2007} based
upon the way bees communicate sources of nectar; and Particle
Swarm Optimisation (PSO) \citep{eberhart:1995} which models the
complex interactions between swarms of insects. These algorithms
can be grouped under the term \emph{swarm intelligence} through
their use of hundreds or thousands of simulated digital organisms.

However, the types of problems that are tackled in nature by these
organisms such as finding sources of food can be considered much
more simplistic than the complex societal problems facing the
human world. In an increasing digital world whereby the available
data is growing considerably alongside inter-connectivity and
joined up thinking, the size and complexity of the problems that
require solving are increasing rapidly such as with \emph{smart
city} planning \citep{batty:2013,murgante:2015}. Moreover, unlike
the natural world, restrictions exist on modern computers in terms
of compute capability and available memory to be able to simulate
many thousands of collective organisms.

In regards to the literature of swarm algorithms most
implementations of collective behaviour algorithms are applied to
relatively small problem sizes. However, there have been some
works in the field addressing scalability. For instance,
\cite{Piccand:2008} found applying PSO to problems of greater than
300 dimensions resulted in failing to find the optimal solution
more than 50\% of the time. \cite{cheng:2015} noted that PSO fails
to scale well to problems of a high dimensionality potentially as
a result of problem structure. However, the authors employ a
social learning implementation whereby many particles act as
demonstrators and present promising results on problems of sizes
up to 1,000 dimensions. \cite{cheng:2015competitive} later propose
a modification to PSO whereby instead of using local and global
best solution to update particle positions a pairwise competition
is performed with the loser learning from the winner to update
their position. The technique demonstrated improved results over
PSO on benchmark problems of up to 5,000 dimensions although it
was noted this was very computationally expensive. \cite{cai:2015}
applies greedy discrete PSO to social network clustering problems
with as many as 11,000 variables. For further reading
\cite{yan:2018} provide a review of the challenges of large-scale
PSO.

Regarding ACO, \cite{li:2011} noted the scaling issues of the
approach proposing a DBSCAN clustering approach to decompose large
Travelling Salesman Problems (TSPs) of up to 1,400 cities into
smaller sub-TSPs and solve these. \cite{ismkhan:2017} also noted
the computational cost and memory requirements and considered the
use of additional heuristics or strategies to facilitate the
scaling of the technique to larger problems. Improvements such as
considering the pheromone matrix as a sparse matrix and using
pheromone in a local search operator enabled ACO to be applied
effectively to TSPs of over 18,000 cities. \cite{chitty:2017} also
noted computational issues with ACO and mitigated them with a
non-pheromone matrix ACO approach which only made partial changes
to good solutions applying the technique to TSP instances of up to
200,000 cities.

Therefore, it can be ascertained both ACO and PSO have issues in
terms of scaling to high dimensional problems, the curse of
dimensionality. Consequently, the question explored in this paper
is can nature inspired, collective intelligence techniques scale
up to the size and complexity of problems that the modern world
desires solving? If not, what are the potential limiting causal
factors for this and what mitigating steps could be taken? These
questions will be investigated using a case study based on ACO to
provide an illustration of the problems faced in scaling up a
collective behaviour meta-heuristic and the hypothesized causal
limitations by applying to a real-world fleet optimisation problem
with steadily increasing complexity. The second aspect of this
paper will attempt to mitigate ACO for these scalability issues
using the novel Partial-ACO approach and enhance the approach
further to assist scalability.

\section{Ant Colony Optimisation: An Exemplar Case}

A popular swarm based meta-heuristic is based upon the foraging
behaviours of ants and known as Ant Colony Optimisation (ACO)
\citep{dorigo:1997}. Essentially, the algorithm involves simulated
ants moving through a graph $G$ probabilistically visiting
vertices and depositing pheromone as they move. The pheromone an
ant deposits on the edges $E$ of graph $G$ is defined by the
quality of the solution the given ant has generated. Ants
probabilistically decide which vertex to visit next using this
pheromone level deposited on the edges of graph $G$ plus potential
local heuristic information regarding the edges such as the
distance to travel for routing problems. An \emph{evaporation}
effect is used to prevent pheromone levels building up too much
and reaching a state of local optima. Therefore, ACO consists of
two stages, the first \emph{solution construction}, simulating
ants, the second stage \emph{pheromone update}. The solution
construction stage involves $m$ ants constructing complete
solutions to problems. Ants start from a random vertex and
iteratively make probabilistic choices using the \emph{random
proportional rule} as to which vertex to visit next. The
probability of ant $k$ at point $i$ visiting point $j\in N^{k}$ is
defined as:\vspace{-0.5cm}

\begin{eqnarray}
    p_{ij}^{k}=\frac{[\tau_{ij}]^{\alpha}[\eta_{ij}]^{\beta}}{\sum_{l\in
N^{k}}[\tau_{il}]^{\alpha}[\eta_{il}]^{\beta}}
\end{eqnarray} where $[\tau_{il}]$ is the pheromone
level deposited on the edge leading from location $i$ to location
$l$; $[\eta_{il}]$ is the heuristic information from location $i$
to location $l$; $\alpha$ and $\beta$ are tuning parameters
controlling the relative influence of the pheromone deposit $
[\tau_{il}]$ and the heuristic information $[\eta_{il}]$.

Once all ants have completed the solution construction stage,
pheromone levels on the edges $E$ of graph $G$ are updated. First,
evaporation of pheromone levels upon every edge of graph $G$
occurs whereby the level is reduced by a value $\rho$ relative to
the pheromone upon that edge:\vspace{-0.2cm}
\begin{eqnarray}
\tau_{ij}\leftarrow(1-\rho)\tau_{ij}
\end{eqnarray}
where $\rho$ is the \emph{evaporation rate} typically set between
0 and 1. Once this evaporation is completed each ant $k$ will then
deposit pheromone on the edges it has traversed based on the
quality of the solution found:\vspace{-0.2cm}
\begin{eqnarray}
\tau_{ij}\leftarrow\tau_{ij}+\sum_{k=1}^{m}\Delta \tau_{ij}^{k}
\end{eqnarray}
where the pheromone ant $k$ deposits, $\Delta \tau_{ij}^{k}$ is
defined by:
\begin{eqnarray}
\Delta \tau_{ij}^{k}&=&\left\{
\begin{array}{ll}
1/C^{k}, & \mbox{if edge $(i,j)$ belongs to $T^{k}$} \\
0, & \mbox{otherwise} \\
\end{array}
\right.
\end{eqnarray}
where $1/C^{k}$ is the quality of ant $k$'s solution $T^{k}$. This
ensures that better solutions found by an ant result in greater
levels of pheromone being deposited on those edges.

\subsection{Consideration of the Scalability of ACO}

From a computational point of view, implementing an ant inspired
algorithm on computational hardware to solve large-scale problems
suffers from three potential limitations regarding overall
performance. The degree of memory required, the computational
costs of simulating thousands of ants and the sheer intractability
of the problem itself.

\vspace{-0.25cm}
\subsubsection{Memory Requirements}
A key aspect of ACO is the pheromone matrix used to store
pheromone levels on all the edges in the graph $G$. This can
require significant amounts of computing memory. For instance, a
fully connected 100,000 city Travelling Salesman Problem (TSP)
will have ten billion edges in graph $G$. Using a float data type
requiring four bytes of memory will need approximately 37GB of
memory to store the pheromone levels, considerably greater than
available in standard computing platforms. In the natural world
storing pheromone levels is not an issue with an infinite
landscape to store them. A secondary memory requirement arises
from ants only updating the pheromone matrix once all ants have
constructed their solutions necessitating storing these in memory
too. For a 100,000 city TSP a single ant will require 0.38MB of
memory using a four byte integer data type. If the number of ants
equals the number of vertices an additional 37GB of memory would
be required.

An ant inspired algorithm that addresses this memory overhead is
Population-based ACO (P-ACO) \citep{guntsch:2002} whereby the
pheromone matrix is removed with only a population of ant
solutions maintained. From this population, pheromone levels are
reconstructed for the available edges by finding the edges taken
within the population from the current vertex and assigning
pheromone to edges based on the solution quality.

\vspace{-0.25cm}
\subsubsection{Computational Costs}
A second aspect to consider with ACO is the time it will take to
simulate ants through the graph $G$. At each vertex an ant needs
to decide which vertex to next visit. This is performed
probabilistically by looking at the pheromone levels, and possibly
heuristic information, on all available edges. This requires
computing probabilities for all these edges. As an example, take a
100,000 city TSP, at the first vertex an ant will have 99,999
possible edges to take all of which require obtaining
probabilities from. Once an ant has made its choice it moves to
the chosen vertex and once again analyses all available edges, now
99,998. Thus, for the 100,000 city TSP an ant will need to perform
five billion edge comparisons. If a processor is capable of 100
GFLOPS (billion floating point operations per second) and assuming
an edge comparison takes one floating point operation it will
require at least 0.05 seconds to simulate an ant through graph
$G$. If using a population of ants equivalent to the number of
vertices in graph $G$ then to complete one iteration of solution
construction would require nearly 90 minutes of computational
time. For ants in nature, compute time is not an issue since each
ant can act independently although, the actual time it would take
real ants to move through a network of this size would still be
problematic.


The simulation of ants is inherently parallel in nature and
therefore can easily take advantage of parallel computing
resources to alliviate the computational costs. In recent years,
speeding up ACO has focused on utilising Graphical Processor Units
(GPUs) consisting of thousands of SIMD processors.
\cite{delevacq:2013} provide a comparison of differing
parallelisation strategies for $\mathcal{MAX}$-$\mathcal{MIN}$ ACO
on GPUs. \cite{cecilia:2013} reduced the decision making process
of ants using a GPU with an \emph{Independent Roulette} approach
that exploits data parallelism and \cite{dawson:2013} went a step
further introducing a \emph{double spin} ant decision methodology
when using GPUs. These works have provided speedups ranging from
40-80x over a sequential implementation, a considerable
improvement. \cite{peake:2018} used the Intel Xeon Phi and a
vectorized candidate list methodology to achieve a 100 fold
speedup. Candidate lists are an alternative efficiency method of
reducing the computational complexity of ACO whereby ants are
restricted to selecting a subset of the available vertices within
its current neighbourhood. If none of these vertices are available
then the full set are considered as normal.
\cite{gambardella:1996} used this approach to solve TSP instances
whereby speedups were observed but also a reduction in accuracy
due to sub-optimal edges being taken.

\vspace{-0.25cm}
\subsubsection{Problem Intractability}
A final scalability issue with ACO involves the amenability of the
problem under consideration to be tackled by ACO. The key issue is
the probabilistic methodology ACO employs to decide which edge to
take next by utilisng the pheromone levels on the available edges
to influence the probabilities. Computationally, an ant will take
the pheromone level on each edge, and if available multiply by the
heuristic information, and multiply this by a random value between
zero and one. The edge with the largest product is selected as the
next to be traversed.

As an example consider a simple decision point whereby an ant has
two choices available, one being the correct, optimal selection,
the other suboptimal. If the pheromone levels on each edge are
equal then there is a 0.5 probability the ant will take the
optimal edge. However, consider ten independent decision points
each with two possible choices akin to a binary optimisation
problem such as clustering a set of items into two groups.
Probabilistically this is equivalent to ten coin flips. With equal
pheromone on all edges, there is only a $0.5^{10}$ probability of
an ant making the optimal choices, approximately one in a
thousand. Conversely, an ant has a 0.999 probability of a
sub-optimal solution so 1,000 ants would need to traverse graph
$G$ to obtain an optimal solution. For a much larger problem of
100,000 decision points this would be $0.5^{100,000}$ requiring
$10^{30,102}$ ants to find the optimal solution.

Consequently, pheromone levels are there to help guide the ants to
taking the optimal edge. Consider the previous 100,000 decision
point example again but with high levels of pheromone on the edge
to the optimal choice, say 0.99 vs. 0.01 on the suboptimal edge,
then the probability of obtaining the correct solution will be
$0.99^{100,000}$ or approximately $3^{437}$ ants required, still a
significant number. In fact, to get to a manageable number of ant
simulations the pheromone on the optimal edges would need to be of
the order 0.9999 vs. 0.0001 on the suboptimal edge when only
approximately 20,000 ants would need to traverse the network
before an ant probabilistically takes the correct edges at each
decision point. However, this means the pheromone level would need
to be 10,000 times greater on the optimal edge than the suboptimal
edge. Moreover, the pheromone levels would need to build up over
time before reaching these levels.

Hence, it can be observed that applying ACO to ever larger
problems results in increasingly reduced probabilities of optimal
solutions being found unless the pheromone levels become
increasingly stronger on the important edges. Candidate lists, as
covered in the previous section, can reduce the number of
decisions that ants need to make but with a potential error
reducing accuracy and can only use if heuristic information is
available to define the neighbourhood. Of course ants in the
natural world do not have problems of this magnitude to solve and
have the numbers if necessary without any undue computational cost
to consider.

\subsection{An Illustration of the Scalability Issues With ACO}
To highlight the potential drawbacks of ACO it will be tested
against a complex set of real-world fleet optimisation problems of
steadily increasing complexity. These problems have been supplied
by a Birmingham based maintenance company which operates a fleet
of vehicles performing services at customer properties within the
city. Each vehicle starts from a depot and must return when it has
finished servicing customers. Each customer is defined by a
location and a job duration predicting the length of time the job
will take and in some cases, a time window for when the jobs must
be completed. The speed of travel of a vehicle between maintenance
jobs is defined at an average 13kph to account for city traffic.
There is also a hard start time and end time to a given working
day defined as 08:00 and 19:00 hours. Fleet optimisation is
essentially the classic Multiple Depot Vehicle Routing Problem
(MDVRP) \citep{dantzig:1959} with Time Windows (MDVRPTW).

The MDVRP can be formally defined as a complete graph $G=(V,E)$,
whereby $V$ is the vertex set and $E$ is the set of all edges
between vertices in $V$. The vertex set $V$ is further partitioned
into two sets, $V_{c}={V_{1},...,V_{n}}$ representing customers
and $V_{d}={V_{n+1},...V_{n+p}}$ representing depots whereby $n$
is the number of customers and $p$ is the number of depots.
Furthermore, each customer $v_{i} \in V_{c}$ has a service time
associated with it and each vehicle $v_{i} \in V_{d}$ has a fixed
capacity associated with it defining the ability to fulfill
customer service. Each edge in the set $E$ has an associated cost
of traversing it represented by the matrix $c_{ij}$. The problem
is essentially to find the set of vehicle routes such that each
customer is serviced once only, each vehicle starts and finishes
from the same depot, each vehicle does not exceed its capacity to
service customers and the overall cost of the combined routes is
minimised.

\begin{table}[!ht]
\scriptsize \centering \caption{\footnotesize Real-world problem
scenarios supplied by a Birmingham maintenance company described
in terms of the vehicles available, customers to service, the
total predicted service time required and the total travel time
using the company's current scheduling.} \label{tab:ProblemData}
\begin{tabular}{c
                c
                c
                c
                c}
\toprule
 & & & {Total Job} & {Total Fleet} \\
{Problem} & {Vehicles} & {Jobs} & {Servicing} & {Traversal Time}\\
 &  &  & {(hh:mm)} & {(hh:mm)} \\
 \midrule
Week\_1 & 8 & 77 & {47:09} & {31:12} \\
Week\_2 & 8 & 79 & {48:24} & {22:49} \\
Week\_3 & 8 & 81 & {48:33} & {19:54} \\
Fortnight\_1 & 16 & 156 & {95:33} & {54:01} \\
Fortnight\_2 & 16 & 138 & {102:01} & {57:07} \\
Fortnight\_3 & 16 & 160 & {96:57} & {42:43} \\
ThreeWeek\_1 & 24 & 237 & {144:06} & {73:55} \\
ThreeWeek\_2 & 24 & 217 & {150:25} & {79:56} \\
ThreeWeek\_3 & 24 & 219 & {150:34} & {77:01} \\
Month\_1 & 32 & 298 & {198:58} & {99:50} \\
Month\_2 & 32 & 313 & {190:26} & {96:28} \\
SixWeek\_1 & 45 & 437 & {267:47} & {142:46} \\
\bottomrule
\end{tabular}\vspace{-0.5cm}
\end{table}

The worksheet data supplied by the company has been divided into a
series of problems of increasing complexity and size which are
described in Table \ref{tab:ProblemData}. The manner in which the
company assigns customer jobs to vehicles is known \emph{apriori}
enabling a \emph{ground truth} for the optimisation process.
Effectively, the company assigns geographically related jobs to
vehicles based on postcode and then orders them such that the
vehicle performs the job furthest from its depot first and then
works its way back, time windows allowing.

\begin{figure}[!ht]
\centering
\includegraphics[width=0.3\textwidth]{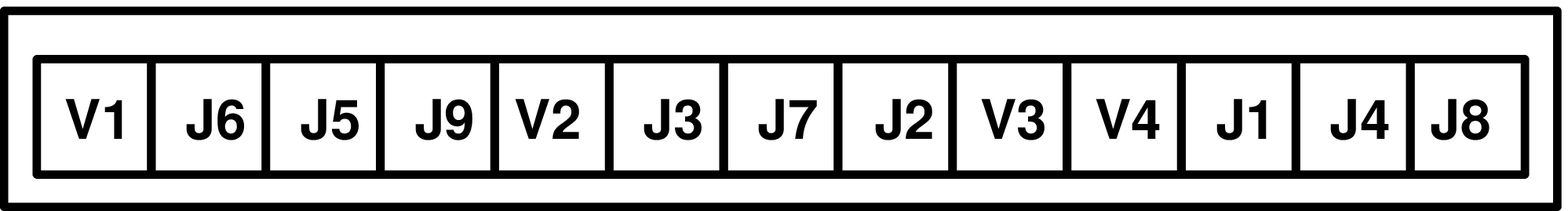}
\scriptsize \vspace{-0.25cm} \caption{\footnotesize Example
solution representation.} \label{fig:Representation}
\end{figure}

To highlight the drawbacks of ACO in terms of scalability, the
$\mathcal{MAX}$-$\mathcal{MIN}$ Ant System ($\mathcal{MM}$AS)
\citep{stutzle:2000} will be tested upon these fleet optimisation
problems. $\mathcal{MM}$AS simulates ants through the graph $G$
but, in contrast to standard ACO, only the best found solution
provides pheromone updates. Additionally, minimum and maximum
levels of pheromone on edges are defined. To solve the fleet
optimisation problem the fully connected graph $G$ has vertices
relating to the number of vehicles and customer jobs. Ants start
from a random vehicle vertex then visit every other vertex once
only resulting in a sequence of vehicles beginning from their
specified depots followed by the customer jobs they will service
before returning to their depot. This representation is shown in
Figure \ref{fig:Representation} whereby \emph{V} relates to a
vehicle and \emph{J} relates to a job. The first vehicle will
undertake jobs 6, 5 and 9, the second jobs 3, 7, and 2 and so
forth.

\vspace{-0.05cm}
\vspace{-0.2cm}

Once a new solution has been generated its quality needs to be
assessed. This is measured using two objectives, the first of
which is to maximise the number of jobs correctly performed within
their given time window. The second objective is the minimisation
of the total traversal time of the fleet of vehicles. Reducing the
number of missed jobs is the primary objective. Hence, comparing
two solutions, if the first services more customer jobs than the
second then the first solution is considered the better. If though
they have equal customer job time serviced then the solution with
the lower fleet traversal time is considered the better.

The pheromone to deposit is calculated using these objectives to
be optimised. A penalty based function will be utilised for the
first objective whereby any customers that have not been serviced
due to capacity limitations or missing the time window will be
penalised by the predicted job time. The secondary objective is to
minimise the time the fleet of vehicles spend traversing the road
network between jobs. Solution quality can then be described as:

\vspace{-0.4cm}
\begin{eqnarray}
C^{k}=(S-s^{k}+1)*L^{k}
\end{eqnarray}
where $S$ is the total amount of time of jobs to be serviced,
$s^{k}$ is the amount of job service time achieved by ant $k$'s
solution and $L^{k}$ is the total traversal time of the fleet of
vehicles of ant $k$'s solution. Clearly, if ant $k$ has achieved
the primary objective of fulfilling all customer demand $S$ then
$C^{k}$ becomes merely the total traversal time of the fleet.

\vspace{-0.4cm}
\begin{table}[!ht]
\scriptsize \centering \caption{\footnotesize Parameters used with
the ACO $\mathcal{MM}$AS algorithm}
\begin{tabular}{p{2cm} p{1cm}}
\toprule
Number of Ants & 192\\
Max Iterations & 1,000,000\\
$\alpha$ & 1.0\\
$\beta$ & 1.0\\
$\rho$ & 0.02\\
\bottomrule
\end{tabular} \centering
\label{tab:params}
\end{table}

\vspace{-0.2cm}

A parallel implementation of $\mathcal{MM}$AS is tested against
the exemplar problems from Table \ref{tab:ProblemData} with
experiments conducted using an AMD Ryzen 2700 processor using 16
parallel threads of execution. The algorithms were compiled using
Microsoft C++. Experiments are averaged over 25 individual
execution runs for each problem with a differing random seed used
in each instance. The parameters used with $\mathcal{MM}$AS are
described in Table \ref{tab:params}.

The results from these experiments are shown in Table
\ref{tab:MMASresults} whereby the issue of scalability is
abundantly clear. As the size and complexity of the fleet
optimisation problems increases, the ability for $\mathcal{MM}$AS
to find a solution which satisfies all the customer demand
reduces. Similarly, $\mathcal{MM}$AS cannot obtain solutions with
a lower fleet traversal time than the company's own scheduling
when the problem size increases. Therefore, it can be considered
that these results support the hypothesis that a nature inspired
swarm algorithm such as ACO suffers from scalability issues.

\vspace{-0.35cm} \sisetup{ table-number-alignment=center,
separate-uncertainty=true, table-figures-integer = 2,
table-figures-decimal = 2}
\begin{table}[!ht]
\scriptsize \centering \caption{\footnotesize The $\mathcal{MM}$AS
results for fleet optimisation in terms of customers serviced,
reductions in fleet traversal time over the original scheduling
and the execution time.}
\begin{tabular}{c
                S[separate-uncertainty,table-figures-uncertainty=1,table-figures-integer = 2, table-figures-decimal = 2, table-column-width=17mm]
                S[separate-uncertainty,table-figures-uncertainty=1,table-figures-integer = 2, table-figures-decimal = 2, table-column-width=18mm]
                S[separate-uncertainty,table-figures-uncertainty=1,table-figures-integer = 2, table-figures-decimal = 2, table-column-width=16mm]}

\hline
{\multirow{2}{1.2cm}{Problem}} & {Job Time} & {Traversal} & {Execution}\\
& {Serviced (\%)} & {Reduction (\%)} & {Time (mins)}\\
\hline
Week\_1 & 100.00 \pm 0.00 & 33.62 \pm 3.39 & 2.23 \pm 0.10 \\
Week\_2 & 100.00 \pm 0.00 & 30.70 \pm 4.85 & 2.33 \pm 0.10 \\
Week\_3 & 100.00 \pm 0.00 & 31.48 \pm 4.68 & 2.49 \pm 0.10 \\
Fortnight\_1 & 100.00 \pm 0.00 & 23.84 \pm 7.46 & 6.56 \pm 0.13 \\
Fortnight\_2 & 100.00 \pm 0.00 & 28.64 \pm 4.99 & 6.84 \pm 0.10 \\
Fortnight\_3 & 100.00 \pm 0.00 & 25.02 \pm 4.49 & 5.76 \pm 0.11 \\
ThreeWeek\_1 & 99.81 \pm 0.18 & -11.43 \pm 7.62 & 13.09 \pm 0.11 \\
ThreeWeek\_2 & 99.95 \pm 0.11 & 7.33 \pm 6.56 & 11.57 \pm 0.15 \\
ThreeWeek\_3 & 99.86 \pm 0.18 & -2.36 \pm 5.92 & 12.09 \pm 0.15 \\
Month\_1 & 99.76 \pm 0.18 & -17.85 \pm 3.75 & 19.75 \pm 0.13 \\
Month\_2 & 99.91 \pm 0.13 & 6.24 \pm 3.47 & 21.57 \pm 0.15 \\
SixWeek\_1 & 98.46 \pm 0.57 & -26.97 \pm 6.76 & 39.54 \pm 0.26 \\
\hline
\end{tabular} \centering
\label{tab:MMASresults}
\end{table}

\vspace{-0.5cm}
\section{Addressing the ACO Scalability Issues}
Given that the evidence seems to support the hypothesis that ACO
methods will struggle to scale to larger, increasingly complex
problems the next step is to attempt to address the underlying
reasons behind the poor performance. As has been previously
discussed, a key problem is the degree of decision making required
to form solutions vs. the probabilistic nature of ACO. Therefore,
it can be theorized that if the degree of decision making is
reduced, ACO may well scale better. A novel modification to the
ACO algorithm known as Partial-ACO \citep{chitty:2017} provides a
mechanism to achieve this. Essentially, this technique minimises
the computational effort required and the probabilistic
fallibility of ACO by ants only considering \emph{partial} changes
to their solutions rather than constructing completely new
solutions. In contrast to standard ACO algorithms, Partial-ACO
operates in a population based manner much the same as P-ACO.
Essentially, a population of ants is maintained each of which
represent a solution to the given problem. Pheromone levels are
constructed from the edges taken within this population of
solutions with their associated qualities which are relative to
the best found solution. Partial-ACO also operates in a pure
steady-state manner to preserve diversity. An ant only replaces
its own \emph{best} solution with a new solution if it is of
better quality. Hence, each ant maintains a \emph{local memory} of
its best yet found solution.

\begin{figure}[!ht]
\centering
\includegraphics[width=0.33\textwidth]{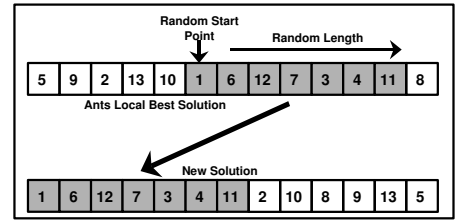}
\scriptsize \vspace{-0.25cm} \caption{\footnotesize An
illustration of the \emph{Partial-ACO} methodology.}
\label{fig:PartialACO}
\end{figure}

This $l_{best}$ memory enables an ant to consequently only
partially change this solution to form a new solution. To
partially modify its locally best found solution an ant simply
picks a random point in the solution as a starting point and a
random sub-length of the tour to preserve. The remaining aspect of
the tour is rebuilt using standard ACO methodologies in a P-ACO
manner. This process is illustrated in Figure
\ref{fig:PartialACO}. To highlight the computational advantage of
this technique, consider retaining 50\% of solutions for a 100,000
TSP problem. In this instance only 50,000 probabilistic decisions
now need to be made and only 1.25 billion pheromone comparisons
would be required, a reduction of 75\%. An overview of the
\emph{Partial-ACO} technique is described in Algorithm
\ref{alg:PartialACO}.

\vspace{-0.5cm}
\begin{algorithm}[!ht]
\footnotesize \caption{\emph{Partial-ACO}} \label{alg:PartialACO}
\begin{algorithmic}[1]
    \FOR{each ant}
        \STATE{Generate an initial solution probabilistically}
    \ENDFOR
   \FOR{number of iterations}
        \FOR{each ant $k$}
            \STATE{Pick uniform random start point from $l_{best}$ solution}
        \STATE{Select uniform random length of $l_{best}$ to preserve}
        \STATE{Copy $l_{best}$ points from start for specified length}
        \STATE{Complete remaining aspect probabilistically}
        \STATE{If new solution better than $l_{best}$ then update $l_{best}$}
    \ENDFOR
\ENDFOR \STATE{Output best $l_{best}$ solution (the $g_{best}$
solution)}
\end{algorithmic}
\end{algorithm}

\subsection{Evaluating the Partial-ACO Approach}

To test the hypothesis that reducing the degree of decision making
that ants need to perform will enable them to scale to larger
problems, Partial-ACO will be tested upon the same problems as
previously. The parameters used for the implementation of
Partial-ACO are described in Table \ref{tab:PartialACOparams}.
Note the lower number of ants in contrast to $\mathcal{MM}$AS. The
original Partial-ACO work found a low number of ants was highly
effective. To ensure the same number of solutions are evaluated,
Partial-ACO will use six times more iterations.

The results for the MDVRP fleet optimisation problem are shown in
Table \ref{tab:PartialACO} whereby it can be observed that now in
all problem instances, reductions in the fleet traversal times are
achieved by Partial-ACO over the commercial company's methodology.
In fact, in many cases the improvement in the reduction in fleet
traversal time is significantly better than that from
$\mathcal{MM}$AS, especially regarding the larger, more complex,
problems. Disappointingly though, the Partial-ACO technique was
also unable to service all the customer jobs for the larger
problems. In terms of execution timings, Partial-ACO is slightly
slower than $\mathcal{MM}$AS when evaluating the same number of
solutions. This is caused by the requirement to construct the edge
pheromone levels at each point as an ant moves through the graph
$G$.

\vspace{-0.3cm}
\begin{table}[!ht]
\scriptsize \centering \caption{\footnotesize Parameters used with
the Partial-ACO algorithm}
\begin{tabular}{p{2cm} p{1cm}}
\toprule
Number of Ants & 32\\
Max Iterations & 6,000,000\\
$\alpha$ & 3.0\\
$\beta$ & 1.0\\
\bottomrule
\end{tabular} \centering
\label{tab:PartialACOparams}
\end{table}

\vspace{-0.5cm}

\sisetup{ table-number-alignment=center,
separate-uncertainty=true, table-figures-integer = 2,
table-figures-decimal = 2}
\begin{table}[!ht]
\scriptsize \centering \caption{\footnotesize The Partial-ACO
results for fleet optimisation in terms of customers serviced,
reductions in fleet traversal time over the original scheduling
and the execution time.}
\begin{tabular}{c
                S[separate-uncertainty,table-figures-uncertainty=1,table-figures-integer = 2, table-figures-decimal = 2, table-column-width=17mm]
                S[separate-uncertainty,table-figures-uncertainty=1,table-figures-integer = 2, table-figures-decimal = 2, table-column-width=18mm]
                S[separate-uncertainty,table-figures-uncertainty=1,table-figures-integer = 2, table-figures-decimal = 2, table-column-width=16mm]}

\hline
{\multirow{2}{1.2cm}{Problem}} & {Job Time} & {Traversal} & {Execution}\\
& {Serviced (\%)} & {Reduction (\%)} & {Time (mins)}\\
\hline
Week\_1 & 100.00 \pm 0.00 & 32.29 \pm 3.77 & 4.69 \pm 0.49 \\
Week\_2 & 100.00 \pm 0.00 & 22.39 \pm 7.84 & 4.76 \pm 0.45 \\
Week\_3 & 100.00 \pm 0.00 & 28.75 \pm 5.55 & 4.90 \pm 0.50 \\
Fortnight\_1 & 99.98 \pm 0.10 & 23.12 \pm 6.30 & 9.87 \pm 0.43 \\
Fortnight\_2 & 100.00 \pm 0.00 & 27.27 \pm 5.49 & 10.18 \pm 0.38 \\
Fortnight\_3 & 100.00 \pm 0.00 & 29.70 \pm 6.64 & 9.09 \pm 0.48 \\
ThreeWeek\_1 & 100.00 \pm 0.00 & 17.13 \pm 4.47 & 16.42 \pm 0.71 \\
ThreeWeek\_2 & 99.91 \pm 0.21 & 20.64 \pm 2.36 & 14.54 \pm 0.78 \\
ThreeWeek\_3 & 99.84 \pm 0.27 & 18.00 \pm 7.84 & 14.95 \pm 0.49 \\
Month\_1 & 99.82 \pm 0.18 & 19.60 \pm 4.09 & 23.59 \pm 0.83 \\
Month\_2 & 99.81 \pm 0.22 & 20.25 \pm 3.60 & 24.73 \pm 1.08 \\
SixWeek\_1 & 97.48 \pm 0.66 & 11.50 \pm 5.78 & 41.71 \pm 0.47 \\
\hline
\end{tabular} \centering
\label{tab:PartialACO}
\end{table}
\vspace{-0.5cm}

\section{Enhancing Partial-ACO}
Although the results of the Partial-ACO approach seemed promising
they did not significantly enforce the premise that ants are less
effective with higher degrees of decision making. Analysing the
Partial-ACO methodology, it could be postulated that modifying a
continuous subsection of an ant's locally best found tour could
present problems in that individual points within the solution
cannot be displaced a great distance. They are confined to a local
neighbourhood as to how they could be reorganised.

\begin{figure}[!ht]
\centering
{\setlength{\fboxrule}{1pt}\fbox{\includegraphics[width=0.33\textwidth]{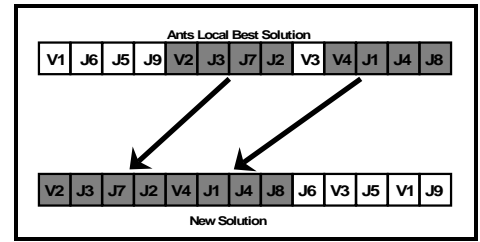}}}
\scriptsize \vspace{-0.25cm} \caption{\footnotesize An
illustration of the \emph{Enhanced} Partial-ACO methodology
whereby two vehicle schedules are preserved.}
\label{fig:EnhancedPartialACO}
\end{figure}

An enhancement to Partial-ACO is proposed which will facilitate
the movement of points in a given ant's locally best solution. To
achieve this, it is proposed that instead of one continuous
segment of an ant's solution being preserved and the remaining
part probabilistically regenerated as is the norm, a number of
separate blocks throughout the solution are preserved instead. In
this way a point at one end of a given solution could be moved to
points throughout the solution. This should help prevent the ants
becoming trapped in local optima. This methodology is actually
well suited to the fleet optimisation problem since each vehicle
can be considered as a stand alone aspect of the solution. Each
preserved block could in fact be a vehicle's complete job
schedule. Before attempting to construct a new solution an ant can
simply decide randomly which vehicle schedules to preserve and
then use the probabilistic behaviour of moving through the graph
$G$ to assign the remaining customer jobs to the remaining
vehicles as normally. Figure \ref{fig:EnhancedPartialACO}
demonstrates the principle whereby it can be observed that two
sections representing vehicle schedules are preserved by an ant
from its $l_{best}$ solution with the rest built up
probabilistically.

\vspace{-0.25cm} \sisetup{ table-number-alignment=center,
separate-uncertainty=true, table-figures-integer = 2,
table-figures-decimal = 2}
\begin{table}[!ht]
\scriptsize \centering \caption{\footnotesize The enhanced
Partial-ACO results for fleet optimisation regards customers
serviced, reductions in fleet traversal time over the original
scheduling and the execution time.}
\begin{tabular}{c
                S[separate-uncertainty,table-figures-uncertainty=1,table-figures-integer = 2, table-figures-decimal = 2, table-column-width=17mm]
                S[separate-uncertainty,table-figures-uncertainty=1,table-figures-integer = 2, table-figures-decimal = 2, table-column-width=18mm]
                S[separate-uncertainty,table-figures-uncertainty=1,table-figures-integer = 2, table-figures-decimal = 2, table-column-width=16mm]}

\hline
{\multirow{2}{1.2cm}{Problem}} & {Job Time} & {Traversal} & {Execution}\\
& {Serviced (\%)} & {Reduction (\%)} & {Time (mins)}\\
\hline
Week\_1 & 100.00 \pm 0.00 & 34.75 \pm 5.92 & 5.08 \pm 0.86 \\
Week\_2 & 100.00 \pm 0.00 & 38.60 \pm 4.14 & 5.15 \pm 0.81 \\
Week\_3 & 100.00 \pm 0.00 & 36.00 \pm 5.05 & 5.76 \pm 1.00 \\
Fortnight\_1 & 100.00 \pm 0.00 & 49.19 \pm 0.57 & 10.75 \pm 0.34 \\
Fortnight\_2 & 100.00 \pm 0.00 & 50.18 \pm 0.49 & 11.25 \pm 0.22 \\
Fortnight\_3 & 100.00 \pm 0.00 & 47.24 \pm 0.80 & 10.21 \pm 0.15 \\
ThreeWeek\_1 & 100.00 \pm 0.00 & 46.55 \pm 1.40 & 18.78 \pm 0.14 \\
ThreeWeek\_2 & 100.00 \pm 0.00 & 42.61 \pm 0.92 & 17.48 \pm 0.14 \\
ThreeWeek\_3 & 100.00 \pm 0.00 & 44.21 \pm 1.20 & 17.57 \pm 0.16 \\
Month\_1 & 100.00 \pm 0.00 & 34.80 \pm 1.94 & 26.22 \pm 0.35 \\
Month\_2 & 100.00 \pm 0.00 & 36.05 \pm 0.92 & 26.96 \pm 0.20 \\
SixWeek\_1 & 100.00 \pm 0.00 & 10.09 \pm 4.16 & 42.36 \pm 0.17 \\
\hline
\end{tabular} \centering
\label{tab:EnhancedPartialACO}\vspace{-0.25cm}
\end{table}

\addtolength{\tabcolsep}{-4pt}
\begin{figure*}
\centering
\begin{tabular}{cccc}
\includegraphics[width=0.245\textwidth]{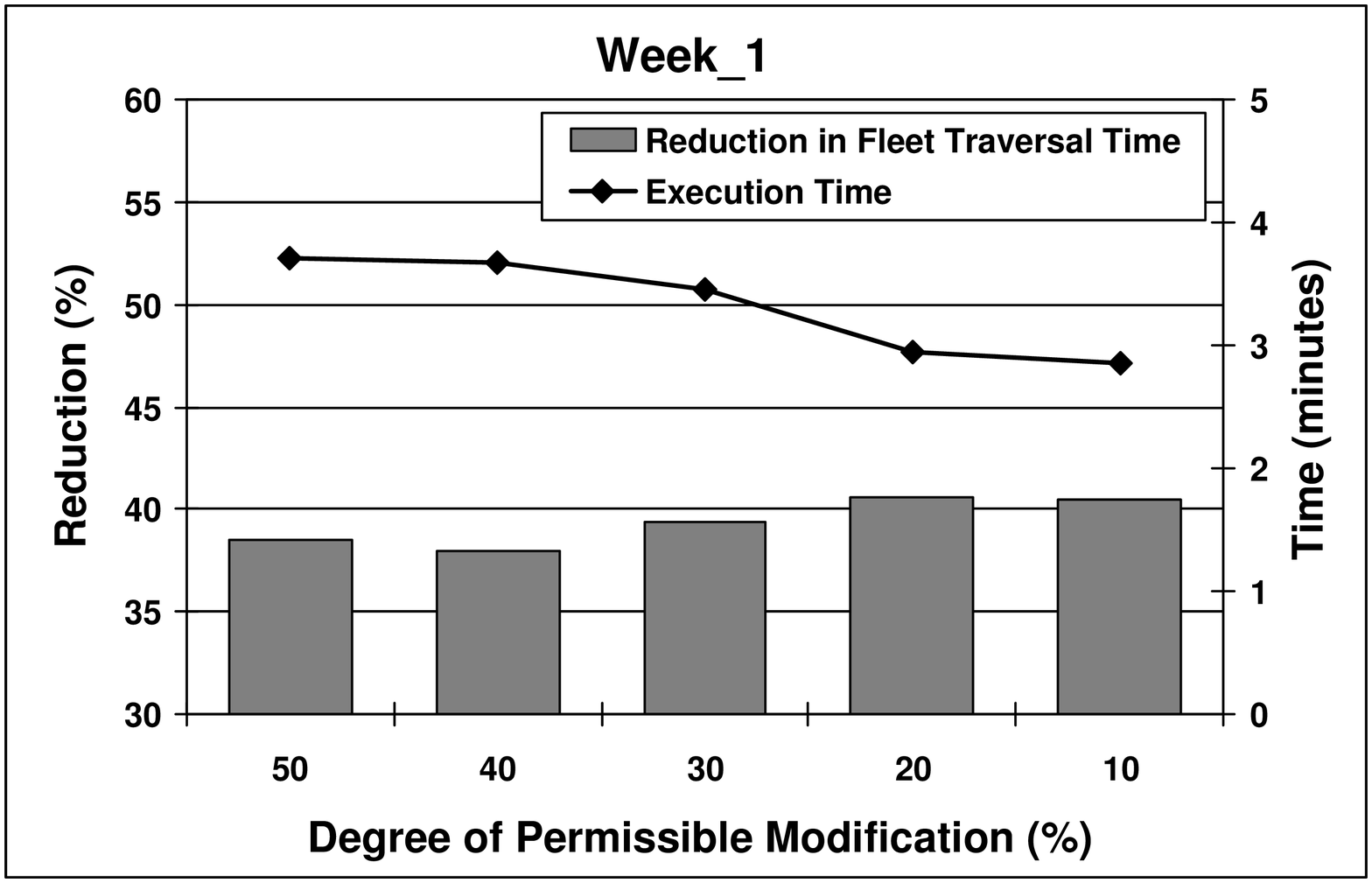} & \includegraphics[width=0.245\textwidth]{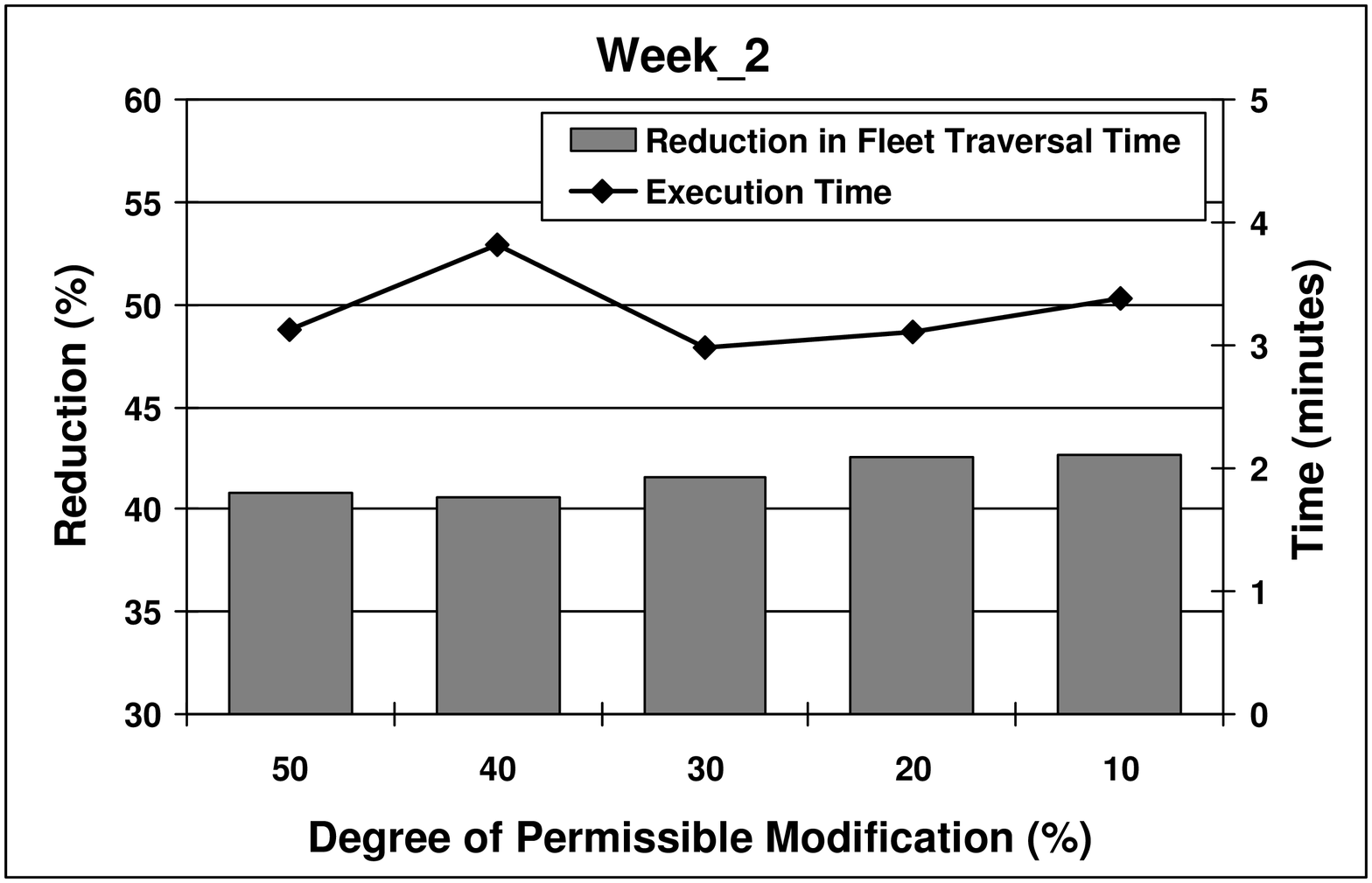} & \includegraphics[width=0.245\textwidth]{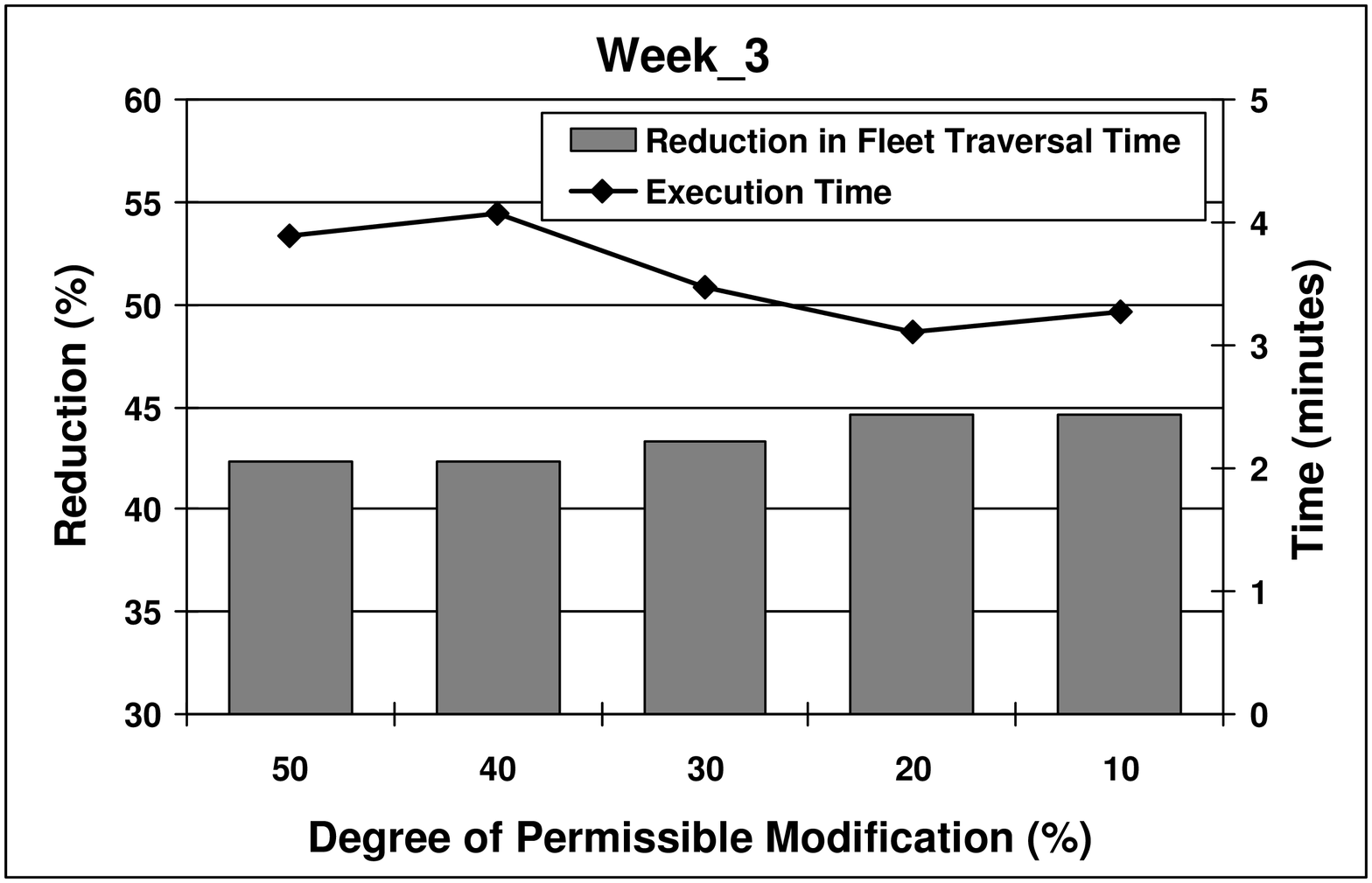} &\includegraphics[width=0.245\textwidth]{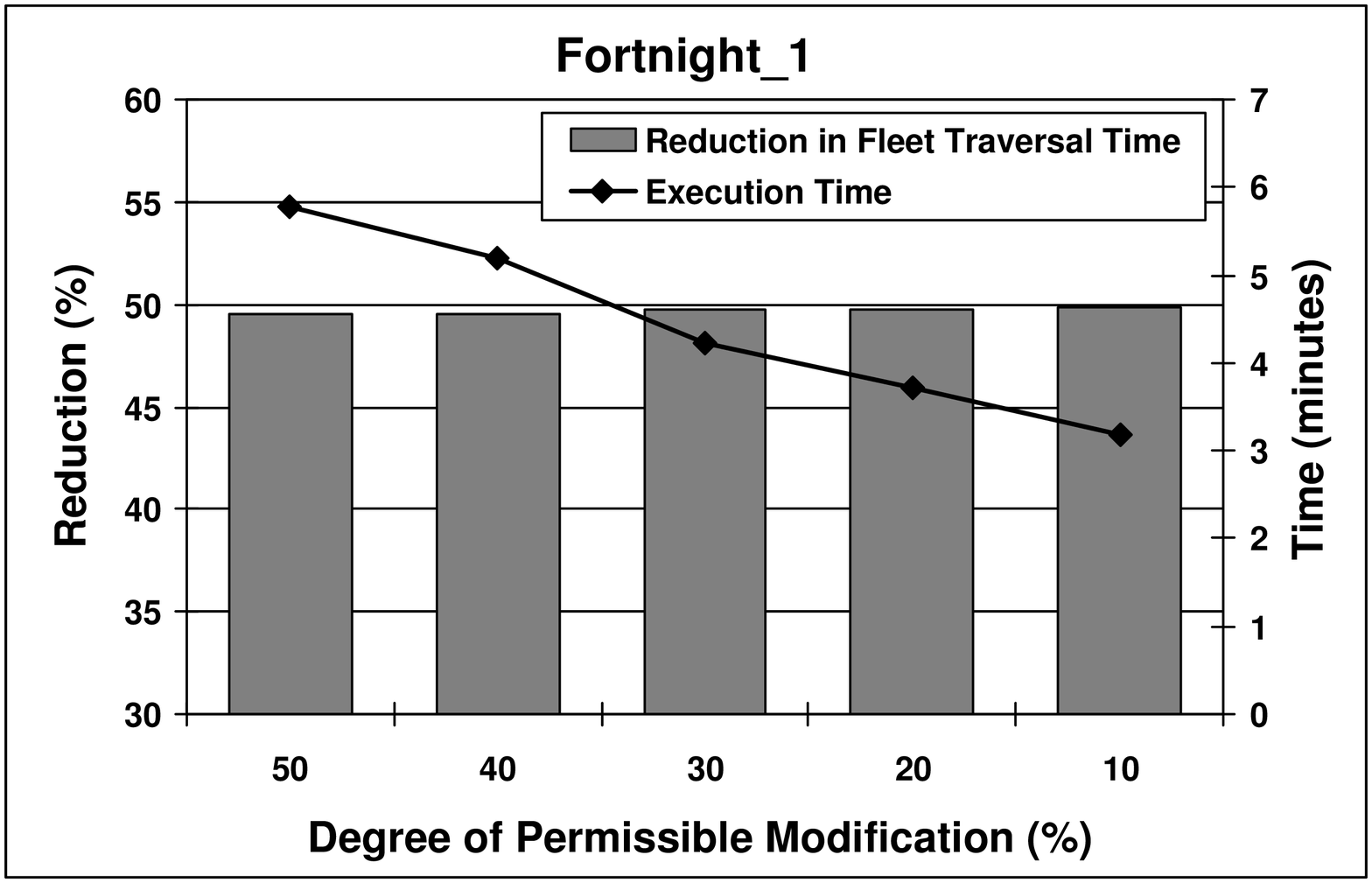}\\
\includegraphics[width=0.245\textwidth]{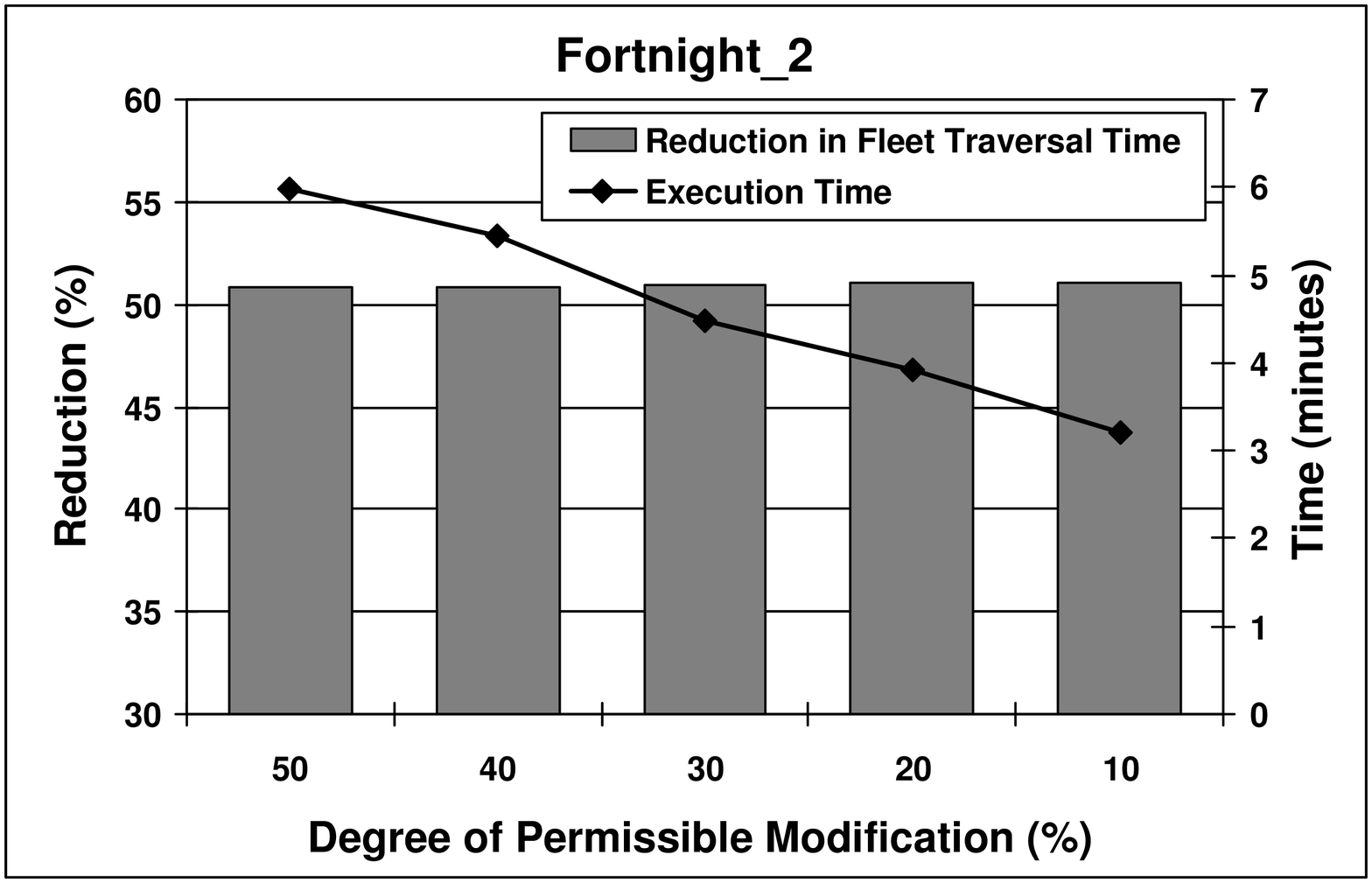} & \includegraphics[width=0.245\textwidth]{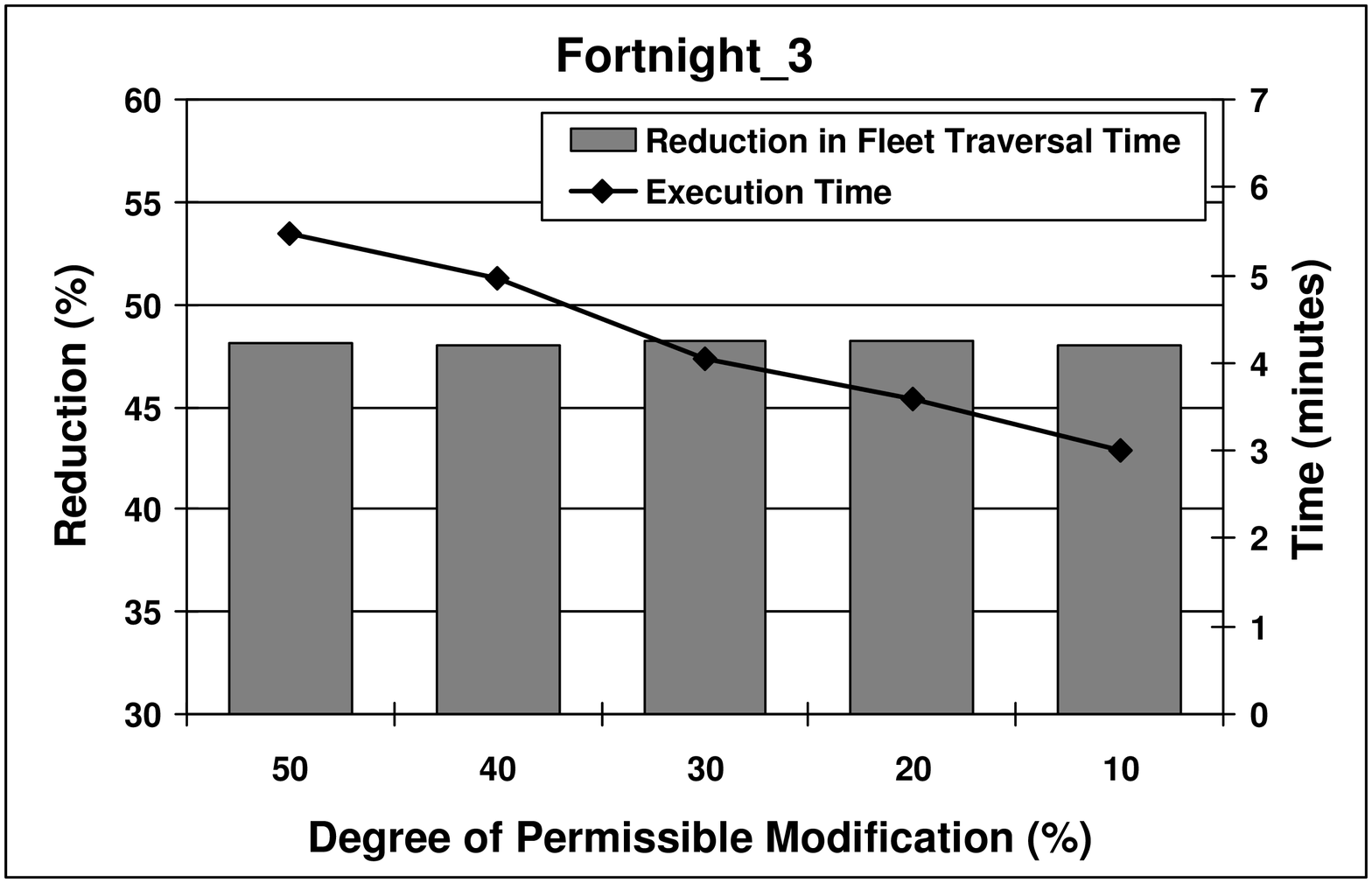} & \includegraphics[width=0.245\textwidth]{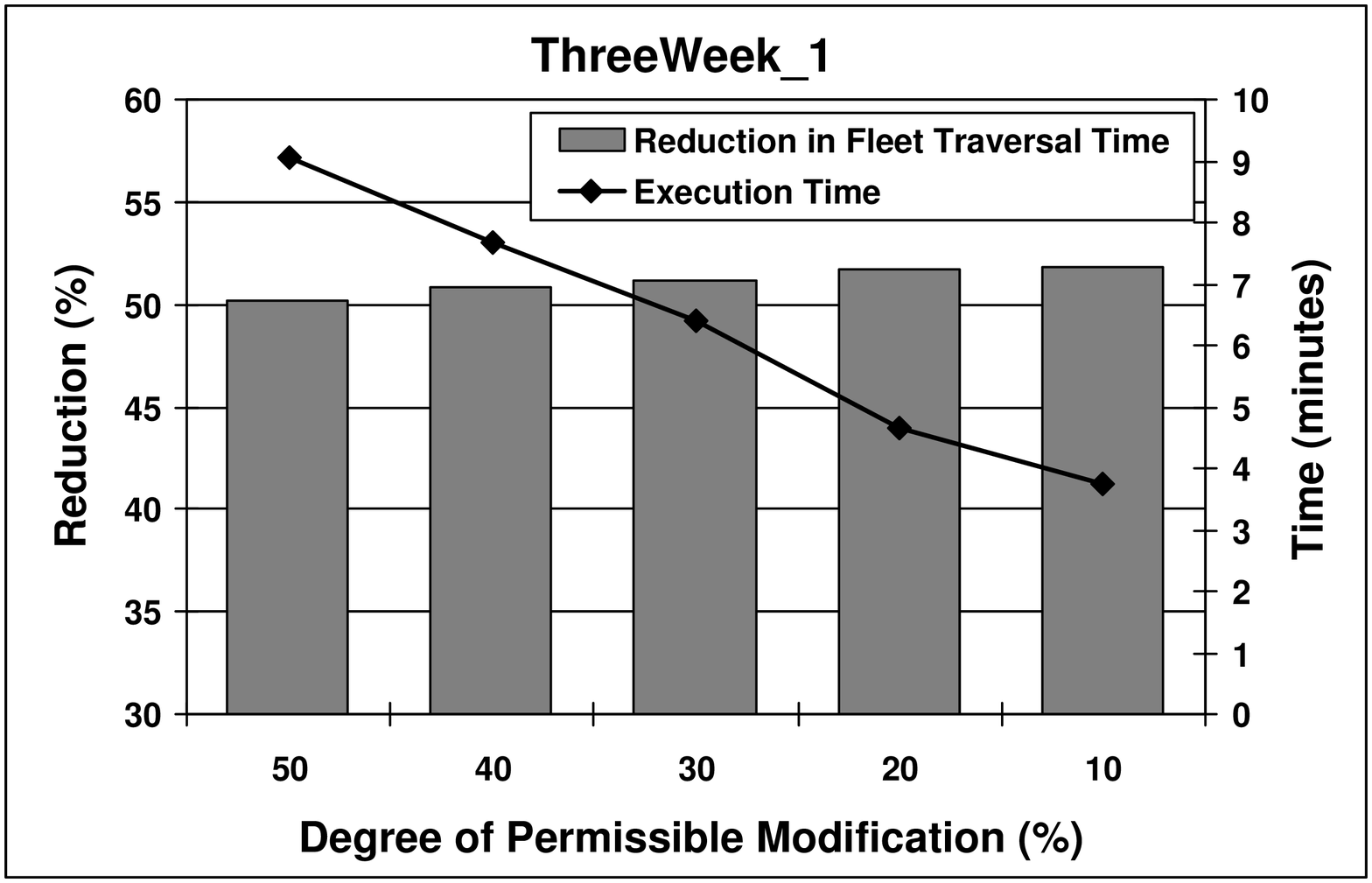} & \includegraphics[width=0.245\textwidth]{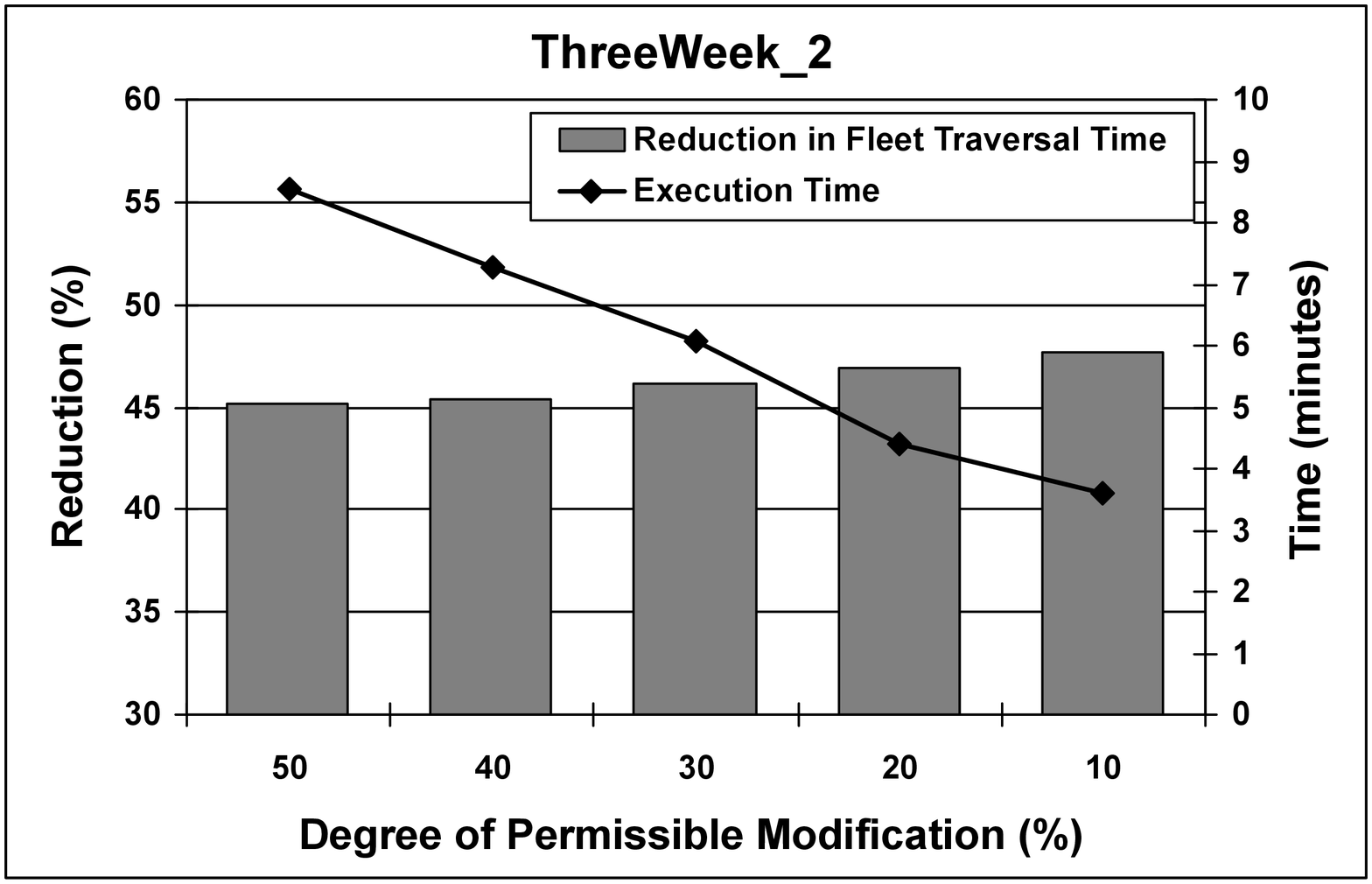}\\
\includegraphics[width=0.245\textwidth]{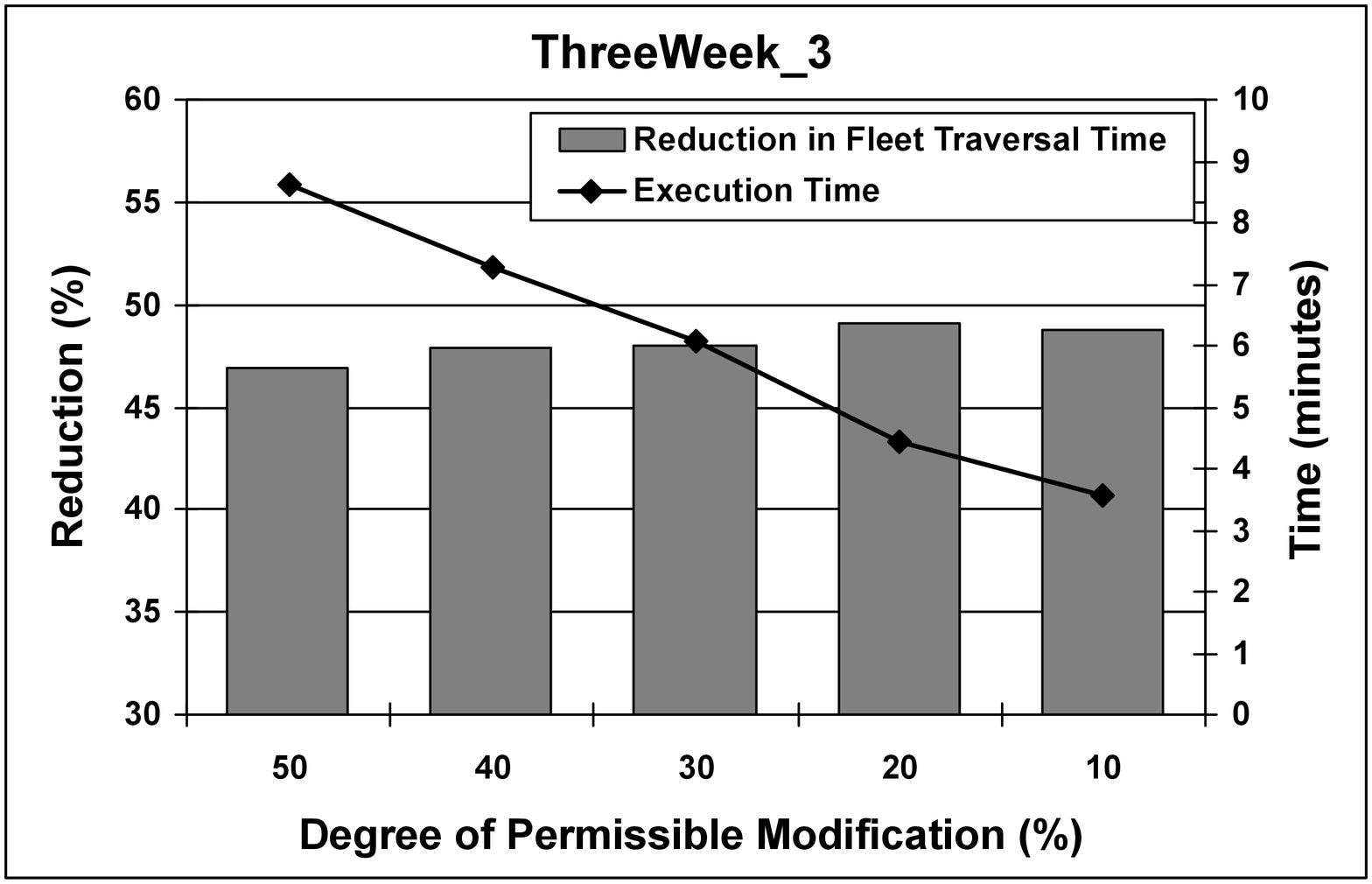} & \includegraphics[width=0.245\textwidth]{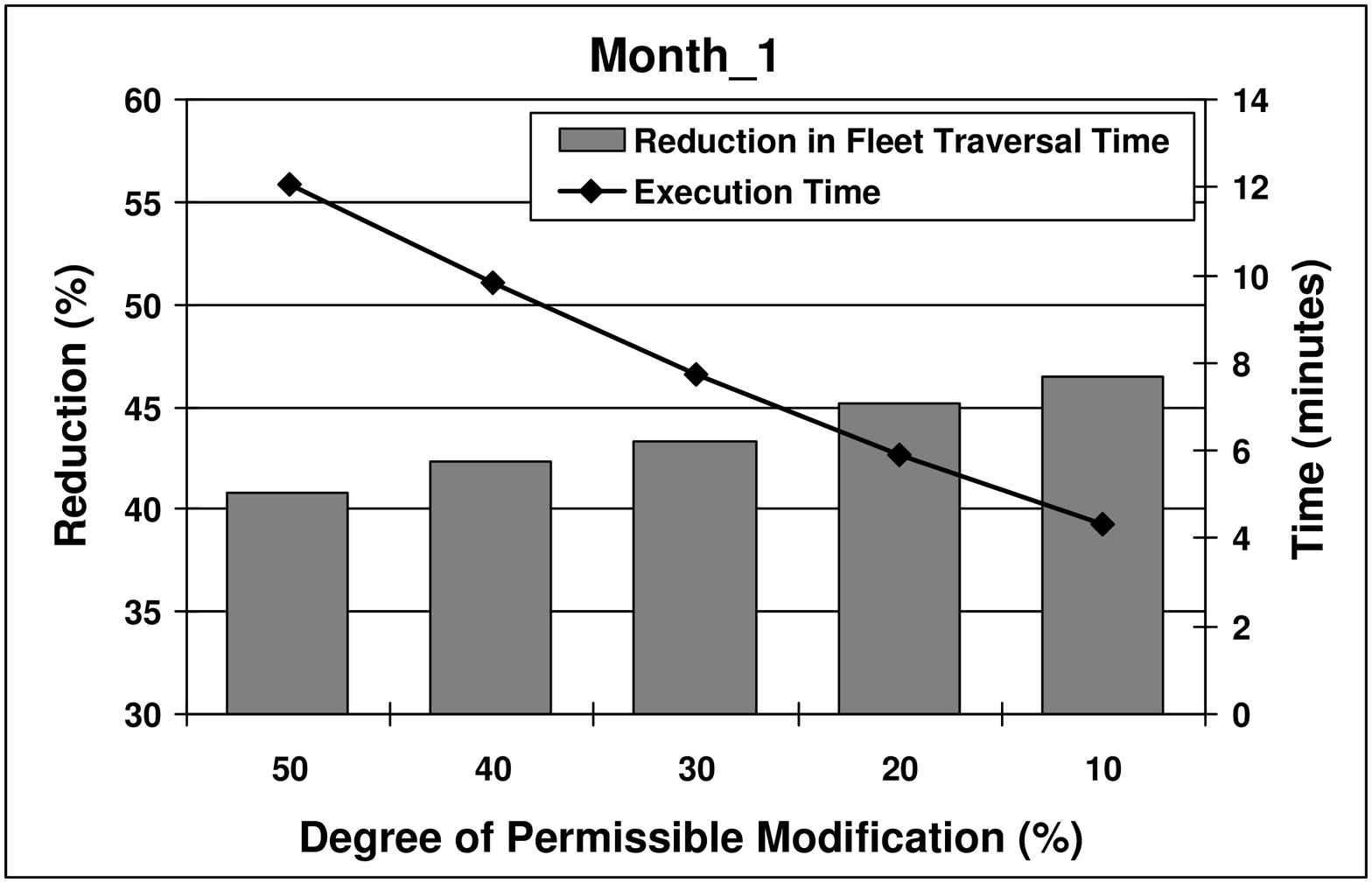} & \includegraphics[width=0.245\textwidth]{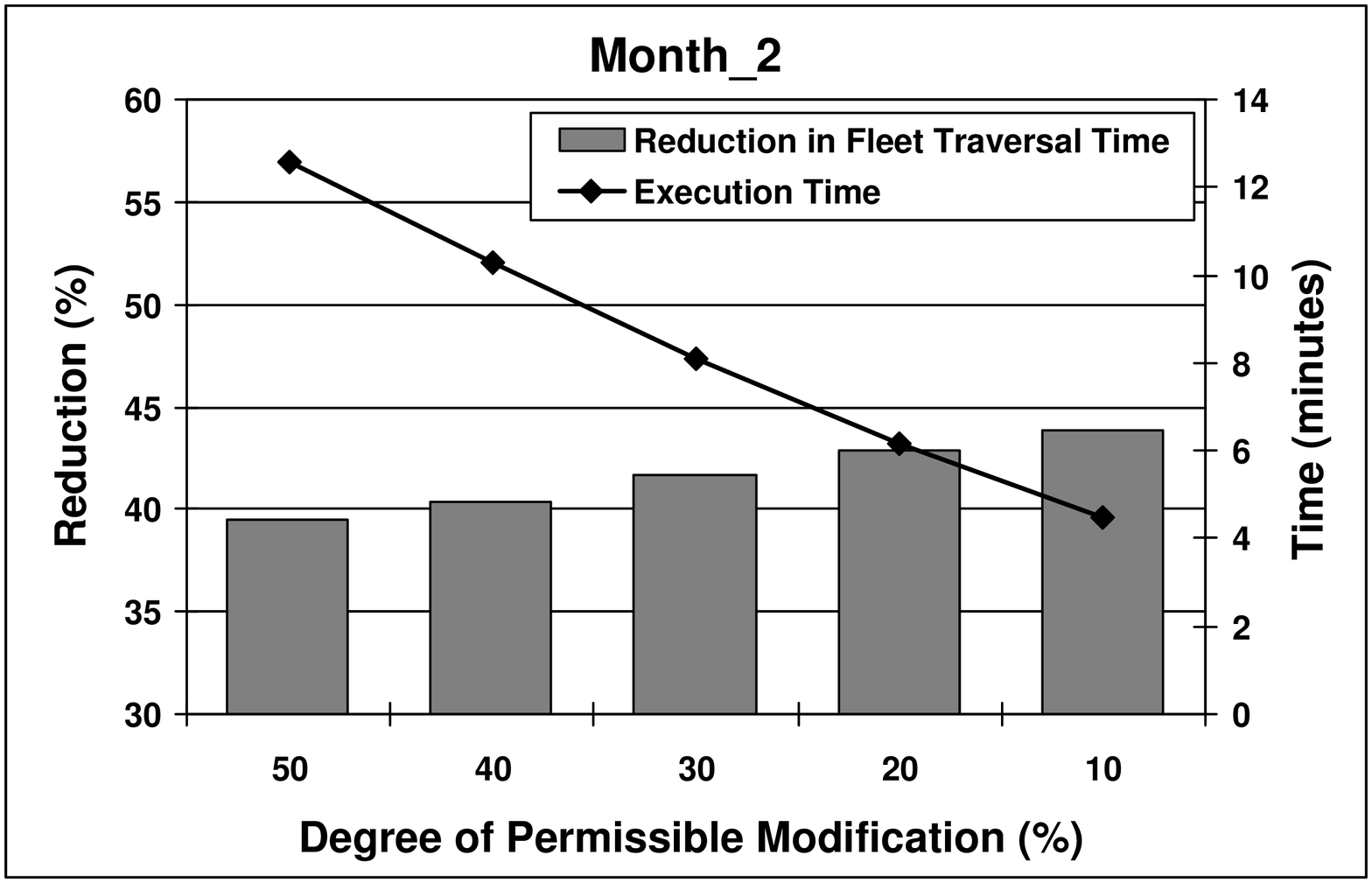} & \includegraphics[width=0.245\textwidth]{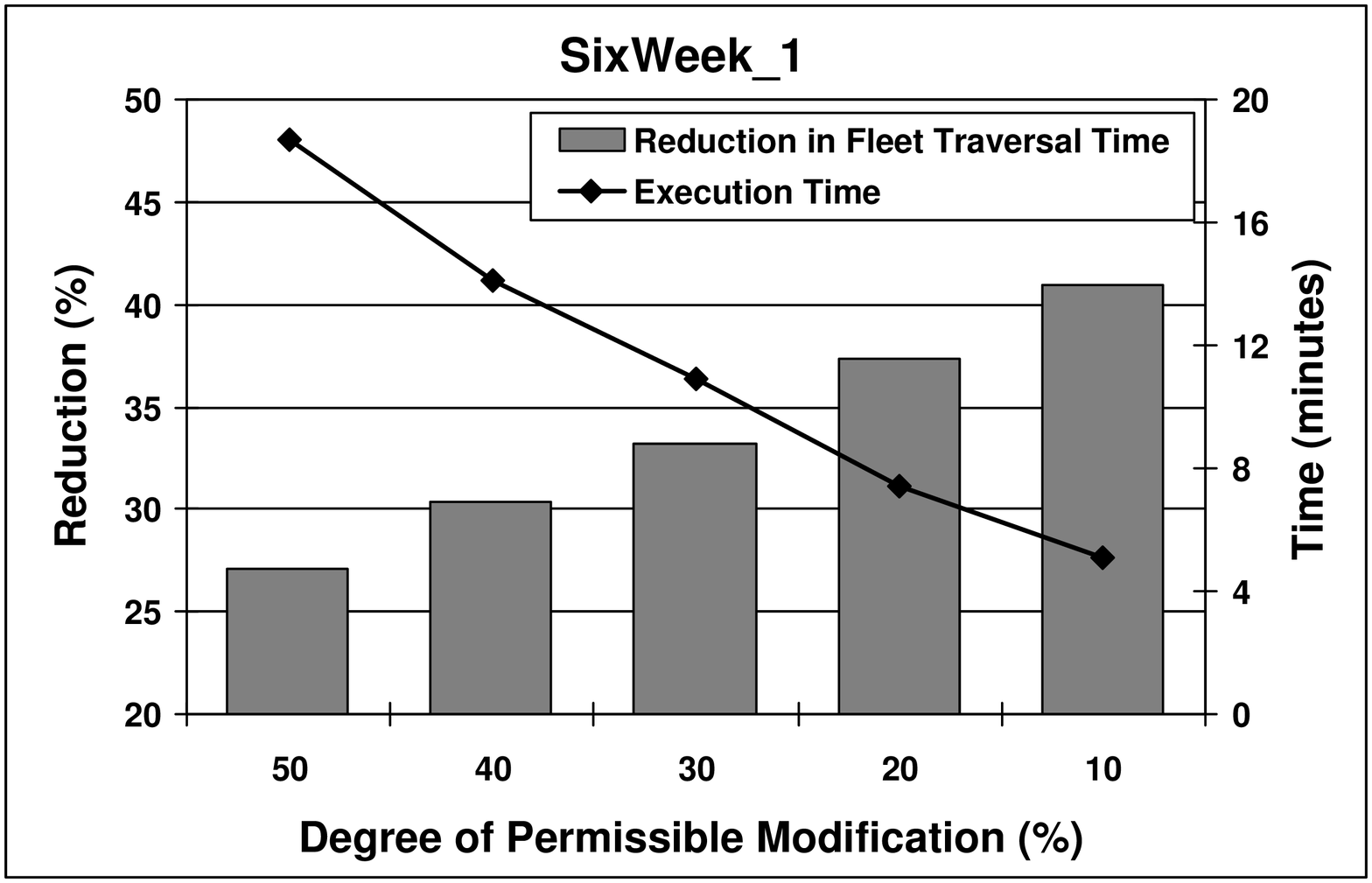}\\
\end{tabular}
\footnotesize \vspace{-0.25cm} \caption{\footnotesize The average
reductions in fleet traversal timings over the commercial
company's methodology as the degree of permissible ant
modification is reduced for each problem instance. Execution
timings are also displayed on the same graph.}
\label{fig:ReducingPermissibleModification}\vspace{-0.5cm}
\end{figure*}
\addtolength{\tabcolsep}{4pt}

To evaluate the enhancement to Partial-ACO, it will be tested
against the same problems as previously using the same parameters
as described in Table \ref{tab:PartialACOparams}. The results are
shown in Table \ref{tab:EnhancedPartialACO} and when contrasted to
those in Table \ref{tab:PartialACO} it can be seen that
significant improvements have been made over the standard
Partial-ACO approach. Now, for all problem instances including the
most complex, all the customer jobs have all been serviced.
Furthermore, significantly improved reductions in the fleet
traversal times have been achieved. In fact, as much as an
additional 26\% reduction in fleet traversal time for the
\emph{ThreeWeek\_2} problem instance. With regards execution
timings, block based Partial-ACO is slightly slower which is
caused by the overhead of assembling blocks of retained solution
rather than one continuous section. Consequently, from these
results it can be inferred that when using solution preservation
with Partial-ACO, smaller random blocks should be preserved rather
than a continuous section to obtain improved results.

\vspace{-0.1cm}
\subsection{Reducing the Degree of Modification}
The enhanced Partial-ACO approach has provided a significant
improvement over standard ACO techniques such as $\mathcal{MM}$AS.
However, recall that the original hypothesis supporting the
development of Partial-ACO was that potentially, collectively
intelligent meta-heuristics could fail to scale well to larger
problems because of the degree of decision making that is
necessitated. This hypothesis seems to be borne out by the results
achieved by Partial-ACO to some degree. However, it is possible to
test this hypothesis to a greater extent by reducing the degree of
permissible modification an ant can make. Currently, an ant will
randomly preserve any amount of its locally best solution and will
modify the rest using the ACO probabilistic rules, approximately
50\% of the solution on average. To avoid a large aspect of
redesign, a maximum degree of modification could be imposed on an
ant changing its locally best solution. This will firstly have the
benefit of increasing the speed of Partial-ACO but also, if the
hypothesis is correct, lead to improved optimisation. As such, the
previous experiments will be rerun using a maximum modification
limit ranging from 50\% of the solution down to 10\% in increments
of 10\%. The improved block preserving version of Partial-ACO will
be used and additionally, to prevent ants becoming trapped in
local optima with a small random probability (0.001) an ant can
modify its locally best found solution to any degree.

The results from reducing the degree of permissible modification
of ants locally best solutions are shown in Figure
\ref{fig:ReducingPermissibleModification}. These describe the
reductions in fleet traversal times over the commercial company's
own scheduling and execution timings. The percentage of customer
jobs serviced is not shown as in all cases 100\% of jobs were
serviced. A clear trend can be observed for improved reductions in
fleet traversal times whilst reducing the degree of permissible
modification. This further reinforces the hypothesis that due to
the probabilistic nature of ants, the degree of decision they are
exposed to must be reduced in order for the technique to scale.
Moreover, the larger the problem, the more pronounced the effect
as evidenced by the month and six week long problem instances.
Remarkably, for the largest problem, reducing ants decision making
by 90\% yields the best results fully enforcing the hypothesis
that ants significantly benefit from reduced decision making. A
further added benefit from reduced decision making of ants is
faster execution times. Not only does the Partial-ACO approach
provide improved reductions in fleet traversal times but can also
achieve these reductions much faster by reducing the probabilistic
decisions that ants need to make. In fact, from these results, it
can stated that Partial-ACO is more accurate, much faster and more
scalable than standard ACO as a consequence of the reduced
decision making of ants within the algorithm.

\section{Conclusions}

This paper has posed the hypothesis that although algorithms
inspired by the collective behaviours exhibited by natural systems
have been effective for simplistic human level problems, they may
fail to problems of much greater complexity. Evidence supporting
this hypothesis is provided by applying Ant Colony Optimisation
(ACO) to a range of increasingly complex fleet optimisation
problems whereby degrading results are observed as complexity
rises. A theory postulated is that the degree of decision making
required by ants to construct solutions becomes too great. Given a
small probability of an ant choosing poorly at each decision
point, the greater decisions required to construct a solution and
available choices, the greater probability of reduced solution
qualities. Consequently, this paper applies the Partial-ACO
approach to reduce the decision making of ants. Indeed, the
Partial-ACO approach provided much improved results for a complex
fleet optimisation problem enabling ACO to scale to much larger
problems with reductions of over 50\% in traversal times achieved
with the subsequent savings in fuel costs for the given company
and a similarly significant reduction in vehicles emissions and
city traffic.

In fact remarkably, for the larger problems, reducing ants
decision making by up to 90\% yielded the best results.
Consequently, this reinforces the posed hypothesis that for
collective behaviour algorithms to scale effectively, the degree
of decision making should be minimised as much as possible.
However, further studies need to be performed with bio-inspired
algorithms besides ACO such as PSO and ABC and to consider problem
areas other than fleet optimisation to provide better supporting
evidence to the hypothesis posed by this paper.

\section{Acknowledgement} Carried out for the System Analytics for Innovation project,
part-funded by the European Regional Development Fund.
\footnotesize
\bibliographystyle{apalike}
\bibliography{AddressingScaleabilityOfACO} 

\end{document}